\documentclass[10pt,journal,compsoc]{IEEEtran}

\usepackage{amsmath}
\usepackage{amssymb}
\usepackage{color}
\usepackage{graphicx}
\usepackage{soul}

\ifCLASSOPTIONcompsoc
  \usepackage[nocompress]{cite}
\else
  \usepackage{cite}
\fi

\begin{document}

\title{The All-Paths and Cycles Graph Kernel}
\author{Pierre-Louis~Giscard
	and~Richard~C.~Wilson,~\IEEEmembership{Senior~Member,~IEEE,}}
\IEEEcompsocitemizethanks{\IEEEcompsocthanksitem P-L. Giscard and R. C. Wilson are with the Department
of Computer Science, University of York,  Heslington, York, YO10 5GH, United Kingdom. P.-L. Giscard acknowledges financial support from the Royal Commission for the
Exhibition of 1851.\protect\\
E-mails: pierre-louis.giscard@york.ac.uk and richard.wilson@york.ac.uk
}

\markboth{IEEE TRANSACTIONS ON NEURAL NETWORKS AND LEARNING SYSTEMS}%
{Shell \MakeLowercase{\textit{et al.}}: The All-Paths and Cycles Graph Kernel}

\IEEEtitleabstractindextext{%
\begin{abstract}
With the recent rise in the amount of structured data available, there has been considerable
interest in methods for machine learning with graphs. Many of these approaches have been
kernel methods, which focus on measuring the similarity between graphs. These generally
involving measuring the similarity of structural elements such as walks or paths. Borgwardt
and Kriegel \cite{shortestpath} proposed the \emph{all-paths kernel} but emphasized that it is
NP-hard to compute and infeasible in practice, favouring instead the shortest-path kernel.
In this paper, we introduce a new algorithm for computing the all-paths kernel
which is very efficient and enrich it further by including the simple cycles as well. We
demonstrate
how it is feasible even on large datasets to compute all the paths and simple cycles up to a moderate length.
We show how to count labelled paths/simple cycles between vertices of a graph and evaluate a labelled path and simple cycles
kernel. Extensive evaluations on a variety of graph datasets demonstrate that the all-paths and cycles
kernel has superior performance to the shortest-path kernel and state-of-the-art performance
overall.
\end{abstract}

}

\maketitle

\section{Introduction}
\label{Intro}
Kernel methods are very popular methods for pattern recognition and machine learning because they
allow the use of a wide variety of classification and clustering tools through the definition of a suitable
\emph{kernel function} for the type of data at hand. A kernel function is a function $K(A,B)$ between
two objects, $A$ and $B$, which in effect measures their similarity. 
Kernels can be used on a wide variety of data including  structured data such as images,
proteins, chemical structures and complex networks where the dominant representation is a graph.
Graph kernels are one of the main paradigms for learning with graphs. Most such kernels are
of the \emph{convolution} type \cite{haussler} which decompose the structural object into parts
and measure the similarity between those parts. Particular examples include the intersection kernel
\cite{cyclicpatternkernels} and the cross-product kernel. In the cross-product kernel, the objects are
decomposed into sets of parts $X_A$ and $X_B$ which are then compared individually
\begin{equation}
K(A,B)=\sum_{x_i\in X_A}\sum_{x_j\in X_B} K_B(x_i,x_j)
\label{CPK}
\end{equation}
where $K_B(.,.)$ is a base kernel measuring the similarity of parts.

Many types of parts have been proposed for such a decomposition, including walks, cycles and paths.
In \cite{shortestpath}, Borgwardt and Kriegel proposed the all-paths kernel, decomposing the graphs
into the set of all paths between any pair of vertices. They noted that this was an NP-hard decomposition,
and very difficult to compute in practice. Instead, they computed the shortest-path kernel which considers
a single path between each pair with the shortest geodesic length.

In this paper, we revisit the all-paths kernel with a new path-counting algorithm. This algorithm is efficient
enough to count all paths up to a moderate length even in large graphs and for large datasets. We
extend this algorithm to count labelled paths efficiently and evaluate the performance on standard datasets.
This shows that considering all paths is a considerable improvement over the shortest-path kernel
and has state-of-the-art performance while remain quick to evaluate.\\[-.5em]

The outline of this paper is as follows. In section \ref{RelatedWork} we review related works on graph
kernels in general and path-based kernels in particular. In section \ref{PathCounting}, we describe the
path-counting algorithm. Section \ref{Labels} extends this algorithm to count labelled paths. Then,
in section \ref{Results}, we evaluate the performance of the method against standard databases.
Finally, in section \ref{Conclusions}, we draw some conclusions.

\section{Related Work}
\label{RelatedWork}

A number of graph kernels have been proposed in the literature, most of which are based on
the idea of a graph decomposition and which generally employ the cross-product kernel
(Equation \ref{CPK}). 
An example is the \textit{subgraph kernel}, which counts the number of isomorphic subgraphs that can be found in a pair of graphs
\begin{equation}
K_{\textrm{SubG}}(G,H)=\sum_{G'\subseteq G}\sum_{H'\subseteq H}\lambda(G')k_{\simeq}(G',H'),
\end{equation}
where the first and second sums run over all subgraphs of $G$ and $H$, respectively, the function $\lambda(.)$ sends any graph to a positive real weight, and $k_{\simeq}(.,.)$ is the isomorphism kernel, that is $k_{\simeq}(G',H') = 1$ if 
and only if $G'$ and $H'$ are isomorphic. Given that the computation of the subgraph kernel is NP-hard \cite{Gartneretal2003}, Shervashidze \textit{et al.} proposed an easier-to-implement alternative with the \textit{graphlet kernel} \cite{Graphlet}. This kernel is based on counting the occurrences of small induced subgraphs of size $\{3,4,5\}$ in the graphs to be classified. More precisely, one forms a feature vector $\mathbf{v}_G$, the $i$th entry of which is the number of times a certain pre-determined subgraph appears in $G$. The kernel value $k_{\text{Graphlet}}(G,G')$ then follows from the dot-product
\begin{equation}
k_{\text{Graphlet}}(G,G')= \tilde{\mathbf{v}}_G^{\mathrm{T}}\,.\,\tilde{\mathbf{v}}_{G'}.
\end{equation}
where $\tilde{\mathbf{v}}_G=\mathbf{v}_G/\sum_i (\mathbf{v}_G)_i$ is a normalised version of the original feature vector. Originally developed for unlabeled graphs, extensions to labelled and attributed graphs have since then been developed \cite{Wale2008, Kriege2012}.

Further examples of kernels based on subgraph comparisons include the \textit{Neighborhood Subgraph Pairwise Distance Kernel} (NSPDK) \cite{costa2010} and \textit{Tree-structured Pattern Kernel} (TPK) \cite{Ramon03expressivityversus, Mahe2009}. The NSPDK is defined as
\begin{multline}
K_{\text{NSPDK}}(G,H)= \\ \sum_{r}\sum_d
\sum_{A_v, B_u\in R^{-1}_{r,d}(G)\atop A'_{v'}, B'_{u'}\in R^{-1}_{r,d}(H)} k_{\simeq}(A_v,A'_{v'})\,k_{\simeq}(B_u,B'_{u'}),
\end{multline}
where $A_v, B_u\in R^{-1}_{r,d}(G)$ designates pairs of neighbourhood subgraphs of $G$, of radius $r$ and the root nodes of which are at distance $d$ from each other. The kernel thus counts identical pairs of neighborhood subgraphs which can be found in both $G$ and $H$. The TP kernel instead counts identical pairs of subtrees in graphs.

Graph kernels based on walks have also been proposed, the simplest example of which is the random walk kernel \cite{Gartneretal2003} which
counts the number of matching walks between two graphs
\begin{equation}
K_\textrm{RW}(G,H)=\sum_{w_i\in W(G)}\sum_{w_j\in W(H)} \alpha(w_i) \delta(w_i,w_j),
\end{equation}
where $W(.)$ is the set of all walks in a graph, $\delta(.,.)$ is the delta function kernel,
i.e. 1 when the walks match and 0 otherwise, and $\alpha(w_i)$ is a weighting function dependent on the length of the walk. Two walks match if they are the same length or in the case of labelled graphs, if they have the same label sequence. This kernel is of interest because it can be efficiently evaluated in cubic time from the adjacency matrix of the product graph, i.e. $K_\textrm{RW}(G,H)=\mathbf{1}^T(\mathbf{I}-\alpha \mathbf{A}_\times)^{-1}\mathbf{1}$.
A number of suggestions have been made to improve the performance of the random walk kernel,
for example to remove backtracking steps from the walk \cite{mahe,backtrackless} since this eliminates
double-counting of structure.

Another alternative to avoid double-counting structure and increase the discriminative power of the
kernel is to use a decomposition into \emph{paths}. A path is a sequence of edges such that no vertex
is repeated in the path. The All-Paths (AP) kernel was proposed by Borgwardt and Kriegel \cite{shortestpath}:
\begin{equation}
K_\textrm{AP}(G,H)=\sum_{p_i\in P(G)}\sum_{p_j\in P(H)} K_B(p_i,p_j)
\end{equation}
where $P(.)$ is the set of all paths in a graph and $K_B(.,.)$ is a base kernel for paths, typically the
delta-function kernel. Borgwardt and Kriegel define a path to be an edge path,
one which may not repeat edges but may repeat vertices, which is different from our definition but this
has little practical impact. They note that computing all the paths is NP-hard in principle
and difficult in practice,
making this kernel impractical. Instead they adopt the \emph{shortest-path} kernel, where $P(G)$ only
contains the paths which are geodesically the shortest between any pair of vertices. Paths are labelled by
their end-point labels. The shortest-path kernel is efficiently computable and practical to implement
but ignores most of the paths present in a graph. 

The kernel which we implement in this work is an extension of the all-paths kernel which includes the simple cycles as well. Recall that a simple cycle is a cycle $v_1v_2\cdots v_{n} v_1$ such that all $v_{1\leq i\leq n}$ are distinct.  Denoting by $PC(G)$ the set of all paths and simple cycles on $G$, we defined the All-Paths and Cycles (APC) kernel to be
\begin{equation}
K_\textrm{APC}(G,H)=\sum_{\gamma_i\in PC(G)}\sum_{\gamma_j\in PC(H)} K_B(\gamma_i,\gamma_j)
\end{equation}
Just as for the paths, counting simple cycles is a $\#$P-complete problem, and the problem of counting paths and simple cycles of length at most $\ell$ parametrised by $\ell$ is a $\#$W[1]-complete \cite{Flum2002}, so the APC kernel is, in principle, hard to compute.

One interesting feature of these kernels is that, if the delta-function base kernel is used, the kernels admit
a straightforward embedding based on the equivalence classes of paths, cycles or walks. We need only form a vector
with elements corresponding to the equivalence classes and store the number of walks in a particular class
in each element. This gives an embedding $\mathbf{v}(G)$ for each graph with the kernel being evaluated
as $\mathbf{v}(G)^T\mathbf{v}(H)$, as with the graphlet kernel.

The \emph{Weisfeiler-Lehmann graph kernel} (WL) is a popular structural kernel which provides
state-of-the-art performance on many tasks \cite{WLkernel}. It also makes use of local structural
elements in the graph, but these are incorporated using a label refinement process\cite{WLpaper}. Each vertex is represented by an intial label. This label is then refined at each iteration by collecting the set of all the labels from
the neighbourhood to form a new label. This process continues up to some maximum level $h$. Each iteration
effectively draws in contextual information from the surrounding neighbour of the graph. The kernel
is then defined as
\begin{equation}
K_\textrm{WL}(G,H)=\sum_{u\in V_G}\sum_{v\in V_H}\sum_{l=0}^h k_\delta (L_l(u),L_l(v))
\end{equation}
where $L_l(u)$ is the label received by vertex $u$ at level $l$. As with the previous kernels, there is a
direct vector-space embedding. This is into the space of all labels, where the presence of a label at
a particular vertex is indicated by a 1 if is appears and 0 if it does not.

The \emph{Weisfeiler-Lehmann optimal assignment kernel} (WLOA) is a recently proposed improvement to
WL which performs better on many datasets \cite{KriegeOA}. Rather than comparing all vertices to
all other vertices when computing the kernel, it employs an optimal pairwise assignment between the vertices
of the two graphs, and only the comparison of the pairs enters into the kernel computation. However, this
approach requires the construction of a hierarchy of labels which can be time consuming for large datasets.

\section{Simple Paths and Simple Cycles Counting Algorithm}
\label{PathCounting}

We begin with a few technical definitions. A \emph{walk} in a graph
is a sequence of vertices $w=v_1v_2\ldots v_n$ such that every adjacent
pair of vertices is joined by an edge, i.e. $(v_i,v_{i+1})\in E$.
The length of the walk is $|w|=n-1$, the number of edges traversed.

We define a \emph{path} in a graph as a walk in which no vertex
is repeated, $v_i\neq v_j,~\forall i\neq j$ except possibly the
first and last vertices, i.e. $v_1=v_n$ is allowed.
If $v_1=v_n$ then the path is a \emph{simple cycle}.
 This is slightly different
from the definition in \cite{shortestpath} where an edge path is defined,
a sequence of edges such that no edge is repeated although vertices may
be repeated. This makes no qualitative difference to the formulation.\\[-.5em]

Counting the number of paths and simple cycles of a particular type that exist in
a graph is a $\#$P-complete problem, the counting equivalent of NP-hardness. This is because the
problem of determining if a graph with $N$ vertices has a path of length
$N$ is known to be NP-complete (the Hamiltonian path problem). In
practice however algorithms can be developed that are more or less efficient
in counting some of the paths. 

In this paper, we present an algorithm for counting the number
of paths between two vertices in a graph $G$, based on the work
in \cite{enumerating}. Our algorithm is based on a recent result from
algebraic combinatorics relating the numbers of walks and paths on any
(directed) graph. This result provides an explicit formula for the ordinary generating function of the number of paths. Let the number of paths of length
$l$ between vertices $u$ and $v$ be $P_{uv}(l)$. Then
\begin{multline}
\sum_l P_{uv}(l)z^l=\left( \sum_{H\prec_\textrm{Conn} G} (z\mathbf{A}_H)^{|H|-1}
(\mathbf{I}-z\mathbf{A})^{|N(H)|}\right)_{uv}\\ \text{for}~u\neq v
\end{multline}
\begin{equation}
\sum_l P_{uu}(l)z^l=\left( \sum_{H\prec_\textrm{Conn} G} (z\mathbf{A}_H)^{|H|}
(\mathbf{I}-z\mathbf{A})^{|N(H)|}\right)_{uu}
\end{equation}
Here $H$ is a \textit{connected}  induced subgraph of $G$ and $\mathbf{A}_H$ is the adjacency matrix
of $H$ zero-extended to match the size of $G$. $|H|$ is the number of vertices
in $H$ and $|N(H)|$ is the number of neighbours of $H$ in $G$ (the `border' of $H$).
Analytic computation of these series then yields two formulas for
calculating the number of paths
\begin{multline}
P_{uv}(l)=(-1)^{l+1}\sum_{\substack{H\prec_\textrm{conn} G\\|H|\leq l+1\\u,v\in H}}
\begin{pmatrix}|N(H)|\\l+1-|H|\end{pmatrix} (-1)^{|H|}
\left(\mathbf{A}_H^l\right)_{uv}\\ \text{for}~u\neq v
\end{multline}
and the number of simple cycles
\begin{equation}
P_{uu}(l)=(-1)^{l}\sum_{\substack{H\prec_\textrm{conn} G\\|H|\leq l\\u\in H}}
\begin{pmatrix}|N(H)|\\l-|H|\end{pmatrix} (-1)^{|H|}
\left(\mathbf{A}_H^l\right)_{uu}
\end{equation}
The formulas for $P_{uv}$ and $P_{uu}$ involve a sum over all connected induced subgraphs
$H$ of size at most $l+1$ and $l$, respectively. Furthermore these subgraphs must contain both $u$ and $v$ in the case of $P_{uv}$ and $u$ in the case of $P_{uu}$.
Finding the subgraphs is achieved by employing the reverse search algorithm introduced by Avis and Fukuda \cite{avis}. Here, we seek all the paths and simple cycles of length at most $l$ on the graph so that all connected induced subgraphs on $l+1$ vertices are both necessary and sufficient. Since the corresponding matrices $\mathbf{A}_H$ before zero extension are small, finding the subgraphs is the most time computationally expensive sub-routine of the algorithm. As a corollary of these observations, once the subgraphs have been found, we obtain both $P_{uv}(k)$ and $P_{uu}(k)$ for all $k\leq l$ at very nearly the same computational cost as would be necessary to get either one of $P_{uv}(l)$ or $P_{uu}(l)$. 
%
A detailed comparison of the overall time complexity of this algorithm
and other counting algorithms for the same task can be found in \cite{counting}. 
\texttt{Matlab} implementations of the paths and cycles counting algorithms used here are available on the File Exchange \cite{CycleCount}.


\section{Counting Labelled Paths and Cycles}
\label{Labels}
The performance of the APC kernel is generally poor when simply counting unlabelled paths. This is
because there is a generic scaling law for the number of paths/simple cycles of length $\ell\ll |G|$ of the form $\mu^\ell$ for typical graphs and so graphs
are only distinguished by different values of the scaling constant $\mu$, also known as connective constant. However, if labels are available, the
number of paths/simple cycles with a particular label sequence is a rich source of information. For simplicity here
we assume that the vertices are labelled rather than the edges, although the framework can
be extended to edge labels. Since approaches to count labelled paths can count labelled simple cycles and vise-versa, here we discuss only this problem in the case of paths without loss of generality.

A na\"ive approach to counting labelled paths involves separately counting the number of paths with
a particular label sequence. Given a set of path counts for all paths of length $l$ and a set of $k$ possible
labels, this involves extending the paths in $k$ different ways. Ultimately this takes $k^l$ evaluations
of the path-counting algorithm. This is both impractical and not useful as the number of possible label
sequences rises more rapidly than the number of paths even for small values of $k$, leading to zero
counts for almost all labelled paths.

An alternative is to code the path labels into edge weights. Since the core path-counting machinery makes
use of multiplication and addition of matrices, the algorithm works equally well with weighted edges and
we can exploit the hardware multiplication of floating-point numbers to efficiently compound different paths
into a single number. Since real numbers are commutative, we cannot recover the order of the edge labels,
but we can distinguish the number of labels in each path. 

Let $L=\{L_1,L_2,\ldots,L_k\}$ be a set of possible vertex labels and $L(u)$ be
the label of vertex $u$. Further let
$\mathbf{c}=(c_1,c_2,\ldots ,c_k)^T$ where $c_i$ is the number of labels of type $L_i$ which appear in a
path. We call $\mathbf{c}$ a \emph{labelling} of the path and define $l(\mathbf{c})=\sum_i c_i$ as the length of the labelling (i.e. the length of the path).
We associate each of the labels with a numeric code $\{s_1,s_2,\ldots,s_k\}$.

As a formal series, we have
\begin{equation}
\mathbf{P}_{uv}=\sum_{\substack{p\in P\\v_1=u,\,v_l=v}} p
\end{equation}
where $p$ is a formal variable representing a particular path and $\mathbf{P}_{uv}$ is
the matrix entry from the path counting algorithm. In a moment, we will replace $p$ with
a code in order to compute $\mathbf{P}_{uv}$ numerically.
Since we know the identity of the first and last vertex ($u$ and $v$ respectively)
we can directly infer the first and last labels of the path without including them in the code. 
The remainder of
each of the paths $v_2 v_3\ldots v_{l-1}$ uses an encoding on edge weights
in the adjacency matrix $\mathbf{A}$. Here we are interested in utilizing vertex weights, so we use a forward encoding where the weight of an edge is the weight of the preceding vertex, i.e. $\mathbf{A}_{ij}=s_i~\forall j$.
Under this encoding, we obtain
\begin{equation}
\mathbf{P}_{uv}=\sum_{\mathbf{c}\in C} m(\mathbf{c}) \prod_{i=1}^k \left(s_i\right)^{c_i} 
\label{pathcounteqn}
\end{equation}
where $C$ contains all labellings of length $l(\mathbf{c})=l-1$ and $m(\mathbf{c})$
is the number of occurrences of that labelling in the paths between $u$ and $v$.
The length is $l-1$ to account for the fact that we do not include the first label as
it is known to be $L(u)$.
Each path labelling receives a code $\prod_{i=1}^k \left(s_i\right)^{c_i}$.
The goal is to decode the path value $\mathbf{P}_{uv}$ to obtain the path labelling counts and to count paths of the form $[L(u),\mathbf{c},L(v)]$.

Equation \ref{pathcounteqn} is a Diophantine equation which must be solved to find the integer values of $m(\mathbf{c})$. This is
hard in general and in practice must be solved either exactly using a search
algorithm or inexactly leading to approximate solutions for the number of paths.
The solutions may also be ambiguous if there are coding collisions, where 
two path codes have a rational ratio and can contribute the same value for
certain path counts. We therefore found it useful to use irrational codes to
reduce this problem; in the experiments presented later three labels are used with
the codes $1,e$ and $\pi$. We find the labelling counts by search, which is computationally
expensive and can only be used for short paths.

Finally the kernel embedding can be computed as the vector $\mathbf{v}(G)$ with entries
$v_{[L(u),\mathbf{c},L(v)]}$ equal to the total number of paths in the whole graph with labelling
$\mathbf{c}$ and initial and final labels $L(u),L(v)$.

As we noted earlier, for even a small number of labels, the number of possible path labellings rises more quickly than
the number of paths in a typical graph. There may thus be no value in seeking an exact count of the labelling
let alone the exact label sequence. 
An alternative is to group the labellings and seek a grouping which is
easy to decode but does not distinguish all labellings. The power labelling is an example of this; we use weights
$(1,a,a^2,a^3\ldots)$ to encode the labels. The coding of a path 
$\mathbf{c}=(c_1,c_2,\ldots ,c_k)^T$ is then $a^{c_2+2c_3+\ldots+(k-1)c_k}$ and
 paths fall into equivalence classes characterised by a particular value of the parameter $c=c_2+2c_3+\ldots+(k-1)c_k$. For two distinct paths to be indistinguishable in the power labelling scheme, they must have the same length and same class value.
Consider for example a graph with three labels $\{A,B,C\}$,
the labellings of a length 3 path $ABB$ and $AAC$ are the same
with $c_2=2, c_3=0$ and $c_2=0, c_3=1$ respectively.

In this scheme, the largest code is $c_\textrm{max}=a^{(k-1)l}$ where $l$ is the path length. Since
double precision arithmetic contains 52 bits of accuracy, this means that $\log_2 a<52/(k-1)l$ in order to accurately
count the paths. For example, with 3 labels and a path length of 5 we can accommodate $a=32$. However, since we
know the identity of the initial and final labels (since we know that the path begins at $u$ and ends at $v$),
in effect we can evaluate paths of length 7. Each code is separated by a factor $a$ and so we can count
up to $a-1$ paths of a class $c$. The path weight is then

\begin{equation}
\mathbf{P}_{uv}=\sum_{c=0}^{(k-1)l} m(c)\, a^c. 
\label{pathclasseqn}
\end{equation}
The path counts $m(c)$ are found by computing the representation of
$\mathbf{P}_{uv}$ in base $a$ which is essentially trivial. 
Again the kernel embedding can be computed as the vector $\mathbf{v}(G)$ with entries
$v_{[L(u),c,L(v)]}$ equal to the total number of paths in the whole graph with class $c$
and initial and final labels $L(u),L(v)$.

In the next section, we compare these two approaches experimentally.

\section{Experimental Evaluation}
\label{Results}

We have compared the performance of the APC kernel to
a number of state-of-the-art convolution kernels and the
recently reported optimal assignment kernel \cite{KriegeOA}.
We used a variety of widely-reported graph classification datasets
drawn from the fields of bio- and chemo-informatics. The
MUTAG, PTC-MR, NCI1 and NCI109 datasets represent a classification
problem on small molecules represented by graphs. Vertex labels exist
in the form of the type of atom present at a vertex and
there are two classes. ENZYMES represents a classification problem into six classes for a graph representation of proteins. There are three vertex labels corresponding to the local configuration of the protein.

\begin{figure}
\begin{center}
\includegraphics[width=1.0\linewidth]{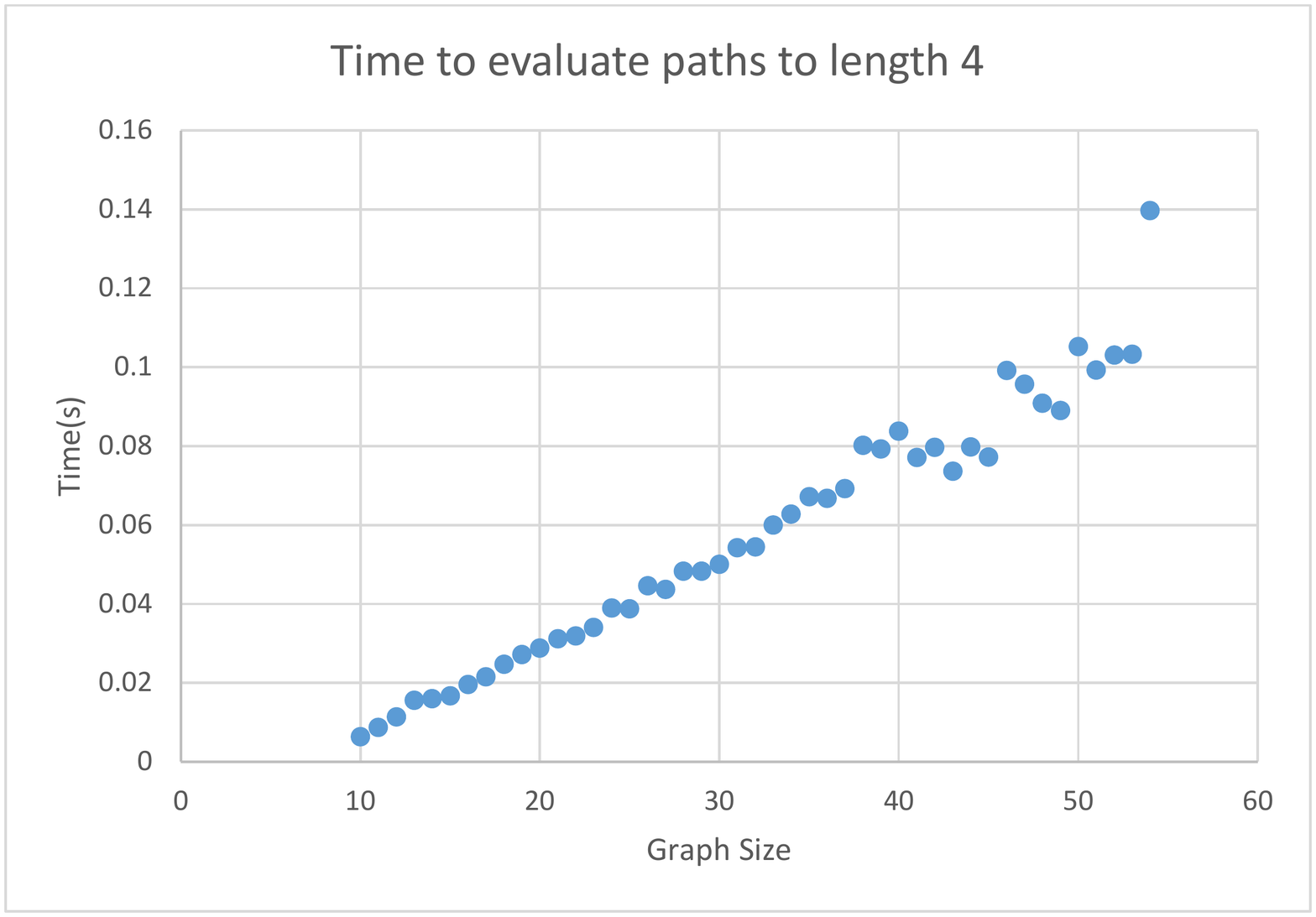}
\caption{The time taken to evaluate all paths of length 6 or less against the size of the graph from dataset NCI1.}
\label{sizetime}
\end{center}
\end{figure}

\begin{figure}
\begin{center}
\includegraphics[width=1.0\linewidth]{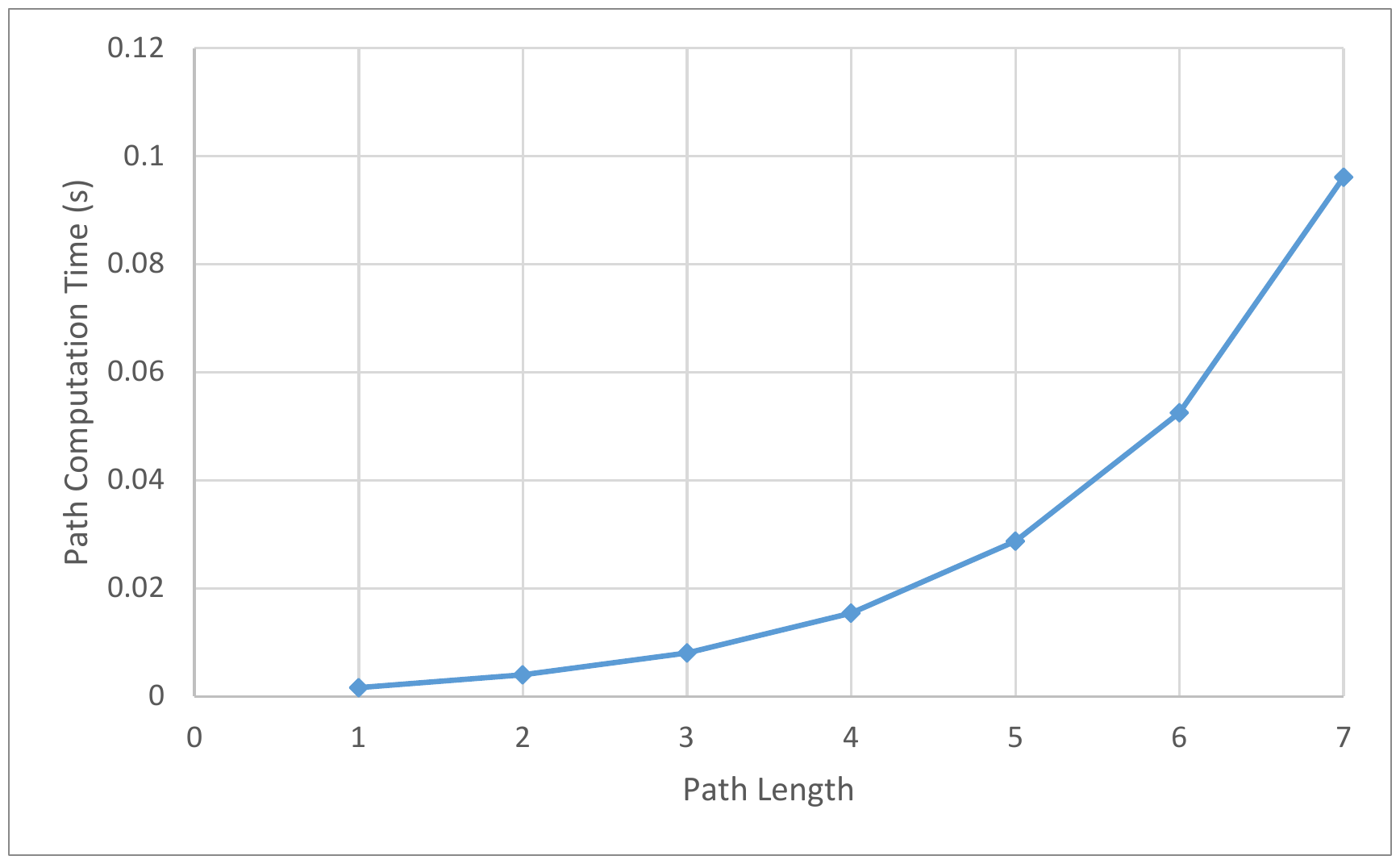}
\caption{The time taken to evaluate all paths in a graph of a particular length against the length for dataset NCI1.}
\label{pctime}
\end{center}
\end{figure}

We begin by examining the computational complexity of the
path finding algorithm in practice. In Figure \ref{sizetime}
we show the time taken against the graph size for evaluating all
paths up to length six on the NCI1 dataset. It is clear that for these graphs, the time taken is linear in the size. Figure \ref{pctime}
shows the relationship between the maximum length of the
paths computed and the time taken. The exponential rise in
computation time is clear, with the time required nearly doubling
for an increment of one in the path length. 
This means that it is not feasible to count very long paths. We give some
specific timings for our datasets later on. Note that both of the scalings with graph size and path length conform with the analysis of \cite{counting}.

In Section \ref{Labels}, we proposed two methods for counting
the number of paths with a particular labelling. The first method
(\emph{exact})
precisely counts the number of paths between two vertices with
a particular labelling $(c_1,c_2,\ldots,c_k)$. The second method
(\emph{power}), which is more efficient to compute, only distinguishes paths with 
different values of the class $c=c_2+2c_3+\ldots+(k-1)c_k$. We now look at
the relative performance of these two methods. Here we use
datasets MUTAG and NCI1-16, which is 
a modified dataset including only graphs of
size 16 or less from the NCI1 dataset, to allow computation to
complete in a reasonable time. The results are shown in Table
\ref{exacttable}.

\begin{table}
\begin{center}
\begin{tabular}{l|cc}
\hline
 & \multicolumn{2}{|c}{\textbf{Dataset}} \\
\cline{2-3}
\textbf{Kernel}	& MUTAG & NCI1-16 \\
\hline
Exact    		& 89.1  & 84.4       \\
Power    		& 89.8  & 83.3       \\
\hline
\end{tabular}
\caption{Classification accuracies(\%) for the two path
labelling methods.}
\label{exacttable}
\end{center}
\end{table}

\begin{figure}
\begin{center}
\includegraphics[width=1.0\linewidth]{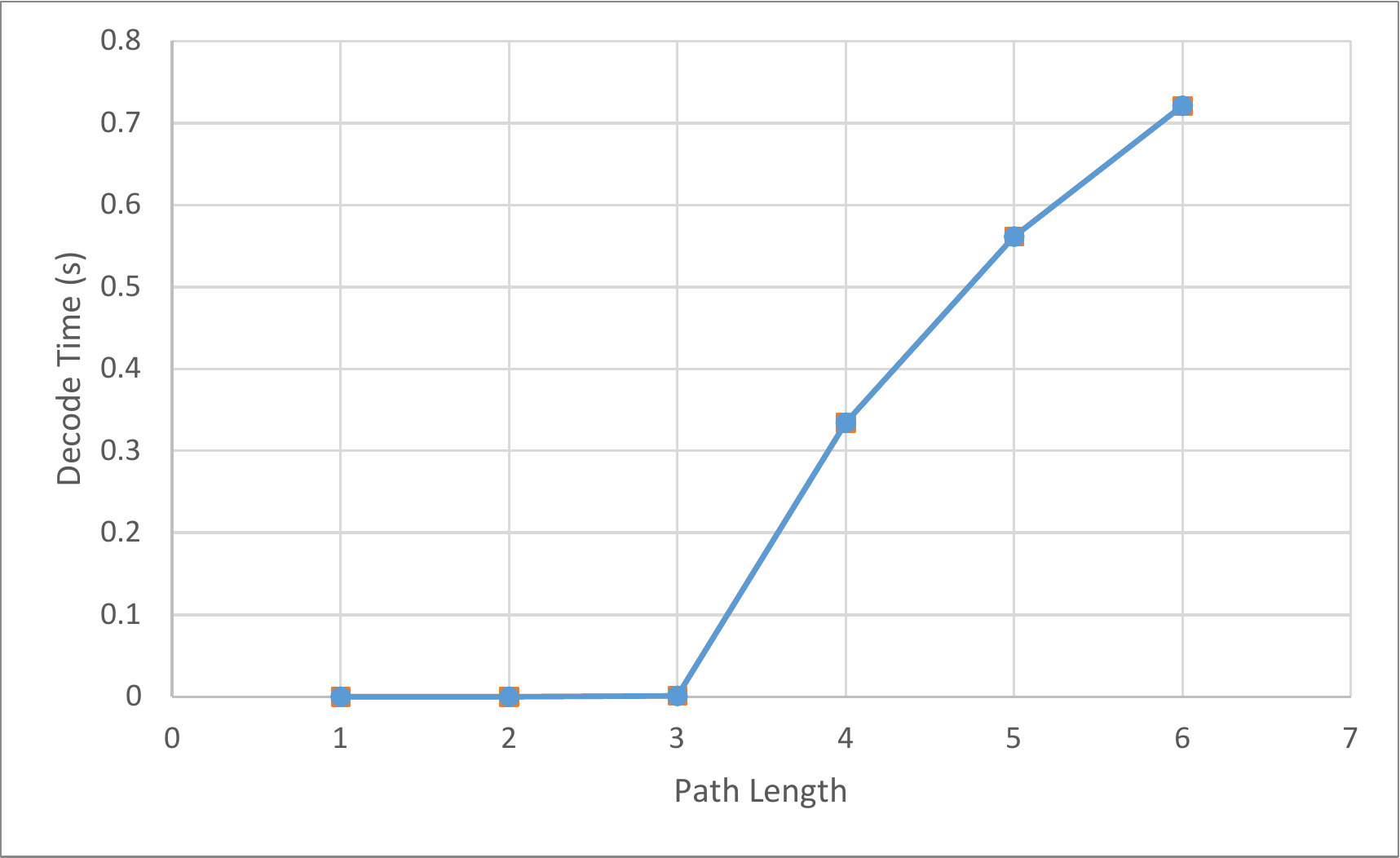}
\caption{The time taken to decode a path value into its
constituent parts using the exact method.}
\label{lengthtime}
\end{center}
\end{figure}

The time taken to do the power decoding is negligible, but
for the exact encoding we must solve the Diophantine equation.
The time taken to do this versus path-length is given
in Figure \ref{lengthtime}. This rises rapidly for paths
of length 4 or longer and becomes the dominant computational
cost. Given the very similar performance of the power encoding,
this latter method is clearly preferable.\\[-.5em]

Finally we test our APC kernel against
four state-of-the-art graph kernels from the literature. 
All Paths  \& Cycles is the method presented in
this paper, using the described algorithms. Path labels are employed
using the power coding using 3 labels. In the datasets MUTAG, PTC-MR, NCI1 and NCI109
the initial labels are mapped onto the two most common labels and
a third representing any other label. ShortestPath is the shortest
path kernel described in \cite{shortestpath}, utilizing the start
and end point labels. The graphlet kernel is the kernel presented earlier in section~\ref{RelatedWork} and first introduced in \cite{Graphlet}.
The Weisfeiler-Lehman subtree kernel (WL) and Weisfeiler-Lehman optimal
assignment kernel (WL-OA) are described in \cite{KriegeOA} and represent
state-of-the-art performance on these datasets.
All of these methods can produce an explicit kernel embedding
and it is this embedding that we use for the sake of efficiency.

Classification is performed using the Matlab implementation of a support vector machine (SVM).
There are a number of adjustable parameters which can be used in
conjunction with the SVM, and these parameters are all set using
cross-validation. We either use no kernel (the raw embedding is used
for classification) or the radial basis function (rbf) kernel.
The embedding co-ordinates may be raw or standardised before use,
and where the rbf kernel is used, it may be auto-scaled. The
parameters used for each method and dataset are listed in Appendix \ref{params}.
10-fold cross-validation is used to compute the classification
performance. The optimal parameters selected are given in
the appendix.

From Table \ref{class} we can see that the APC kernel does indeed give a significant
performance boost over the shortest-path kernel, giving a 5-9\% improvement.
All Paths \& Cycles is competitive across the board, giving superior performance
to Shortest-path and Graphlets, and comparative performance to the state-of-the-art Weisfeiler-Lehmen kernel. The recent WL optimal assignment kernel out-performs
APC, but as we shall see in the next section, this kernel is expensive on large
datasets.

\begin{table*}
\begin{center}
\begin{tabular}{l|ccccc}
\hline
 & \multicolumn{5}{|c}{\textbf{Dataset}} \\
\cline{2-6}
\textbf{Kernel}& MUTAG & PTC-MR & NCI1  & NCI109 & ENZYMES \\
\hline
All Paths \& Cycles       & 89.8  & 62.2   & 80.0  & 79.9   & 53.3	  \\
ShortestPath   & 85.1  & 61.2   & 75.6  & 75.1   & 44.8	  \\ 
\hline
Graphlets\footnotemark
               & 85.2  & 54.7   & 70.5  & 69.3   & 30.6   \\
WL             & \textbf{90.8}  & 63.5   & 85.4  & 85.7   & 50.2   \\
WL-OA          & 85.3  & \textbf{65.1}   & \textbf{86.0}  & \textbf{85.8}   & \textbf{56.6}   \\
\hline
\end{tabular}
\caption{Classification accuracies (\%) on standard graph datasets. The accuracies are $\pm 0.5\%$ except for the Graphlets kernel}
\label{class}
\end{center}
\end{table*}

\begin{table*}
\begin{center}
\begin{tabular}{l|ccccc}
\hline
 & \multicolumn{5}{|c}{\textbf{Dataset}} \\
\cline{2-6}
\textbf{Kernel}& MUTAG & PTC\_MR & NCI1  & NCI109 & ENZYMES \\
\hline
All Paths \& Cycles       & 3.5   & 4.2    & 38.7  & 38.1   & 19.4	  \\
ShortestPath   & 0.5   & 0.4    & 70.3  & 71.6   & 2.4	  \\
WL             & 0.1   & 0.4    & 27.5  & 33.0   & 0.4    \\
WL-OA          & 0.1   & 0.4    & 2196   & 2257    & 16.6   \\
\hline
\end{tabular}
\caption{Computation time (in seconds) on standard graph datasets.}
\label{time}
\end{center}
\end{table*}

The second issue is the time taken to compute these kernels. The total computation
time is given in Table \ref{time}. All methods were implemented in Matlab and
run on an Intel Core i7-4770 3.4GHz processor with 16GB of memory.
The exception is the evaluation of the histogram intersection kernel
necessary for the WL-OA kernel, which was found to be very slow in
Matlab and was replaced by a compiled routine (and so is partially optimized).
Since
all methods could be further optimised, these figures are only indicative.
The methods were run with the parameters used to achieve the performance
in Table \ref{class}.
All Paths \& Cycles can be computed in a very reasonable time for all datasets,
taking a maximum time of 38.7 seconds on the NCI1 dataset which has
4110 graphs of average size of 30 vertices. In comparison to ShortestPath,
it is slower on the smaller datasets MUTAG, PTC\_MR and ENZYMES, but much quicker on the NCI datasets.
The WL kernel is very quick
to compute but has a time-cost which rises rapidly with the size of the dataset, as does the high-performance WL-OA kernel.
There are several reasons for this behavior. For higher numbers of iterations in the WL kernels, the
feature spaces become very large. In NCI1, a depth of 7 is required for good performance which generates some
200,000 unique labels (and therefore features). On the same dataset, APC uses just 81 features.
We anticipate that for larger datasets, APC would be quicker with the same performance. 
WL-OA gives the best classification performance, but the large feature space means that
 the cost of computing the histogram intersection required for the kernel is prohibitive on the
larger datasets.
For the large datasets, the performance of APC is still good and WL-OA cannot
compete in terms of speed. Our data suggests that APC is superior for very large
datasets.
\footnotetext{Results reproduced from \cite{KriegeOA} with permission. Uncertainties are given in that paper}

\section{Conclusions}
\label{Conclusions}
We have presented an implementation of the All Paths \& Cycles kernel. The version of this kernel based only on the simple paths had originally been proposed in 2004 but was immediately dismissed as too difficult to implement owing to the underlying $\#$P-complete problem of counting these paths. Of course, the complexity hierarchy has remained unchanged since then, yet, in practice, the latest algorithm for counting both paths and simple cycles has enough performance for these to be counted on real-world networks. Furthermore, by using well chosen numerical values to represent vertex labels, we have shown here that the path-counting algorithm can count labelled paths at no further computational expense. These breakthroughs effectively make the APC kernel feasible in graph-classification problems. To test the kernel performances, we proceeded at extensive tests on the MUTAG, NCI1, NCI109 and ENZYMES datasets. First, we found the APC kernel to clearly out-perform the shortest-path kernel, an ersatz to the all-paths one originally developed to circumvent computational difficulties. Second, we showed that the APC kernel offers state-of-the-art performances overall, 
equivalent to the best-performing methods on e.g. MUTAG and ENZYMES. In addition the kernel offers excellent computation times on all datasets. In particular on the larger datasets NCI1 and NCI109 the computation time compares favourably, being much quicker to evaluate than, for example the Weisfeiler-Lehman optimal
assignment kernel and the shortest path kernel.

\appendix

\section{Experimental parameter settings}
The kernels generate features to a pre-selected number of levels $L$,
which is the number of refinements for the WL-kernels and the path\/cycle length
for the All Paths \& Cycles kernel.
The raw feature vectors from the embedding of each kernel are
first processed by removing any very low variance features; `cut' gives
the fraction of the total variance below which the feature is discarded.
The features are then classified by a support vector machine (with
error-correcting output codes for ENZYMES which has more than two classes),
the remaining parameters refer to the SVM kernel used, and whether the
features are standardised before use.

\begin{table}[h]
\begin{center}
\begin{tabular}{l|c|ccccc}
\hline
 Dataset & Method & $L$ & Cut & Kernel & K-scale & Stand. \\
\hline
		& APC & 4  & $10^{-5}$ & rbf    &  auto   &	\checkmark	\\
	& SP & -  & $10^{-5}$ &  -     &         &	\checkmark	\\
MUTAG	& WL       & 2  & $10^{-5}$ &  -     &         &		\\
		& WL-OA    & 2  & $10^{-5}$ &  -     &         &		\\
\hline
		& APC & 5  & $10^{-5}$ & rbf    & auto    &	\checkmark	\\
	& SP & -  & $10^{-5}$ &  -     &         &	\checkmark	\\
PTC\_MR	& WL       & 4  & $10^{-5}$ & rbf    & auto    &		\\
		& WL-OA    & 3  & $10^{-5}$ & rbf    & auto    &		\\
\hline
		& APC & 3  & $10^{-5}$ & rbf    & auto    &	\checkmark	\\
	& SP & -  & $10^{-5}$ & rbf    & auto    &	\checkmark	\\
NCI1	& WL       & 7  & $10^{-5}$ & -       &		\\
		& WL-OA    & 7  & $10^{-5}$ & -       &		\\
\hline
		& APC & 3  & $10^{-5}$ & rbf    & auto    &	\checkmark	\\
	& SP & -  & $10^{-5}$ & rbf    & auto    &	\checkmark	\\
NCI109	& WL       & 7  & $10^{-5}$ & -    &     &		\\
		& WL-OA    & 7  & $10^{-5}$ & -    &     &		\\
\hline
		& APC & 3  & $10^{-6}$ & rbf    & auto    &	\checkmark	\\
	& SP & -  & $10^{-5}$ & rbf    & 1       &	\checkmark	\\
ENZYMES & WL       & 2  & $10^{-5}$ &  -     &		&\\
		& WL-OA    & 3  & $10^{-5}$ &  -      &         &		\\
\hline
\end{tabular}
\caption{Parameters used to generate the results in Table \ref{class}.}
\label{params}
\end{center}
\end{table}


\end{document}